\documentclass[
]{ceurart}

\usepackage{listings}
\begin{document}

\copyrightyear{2025}
\copyrightclause{Copyright for this paper by its authors.
  Use permitted under Creative Commons License Attribution 4.0
  International (CC BY 4.0).}

\conference{Challenge and Workshop (BC9): Large Language Models for Clinical and Biomedical NLP, International Joint Conference on Artificial Intelligence (IJCAI), August 16–22, 2025, Montreal, Canada}

\title{NOWJ @BioCreative IX ToxHabits: An Ensemble Deep Learning Approach for Detecting Substance Use and Contextual Information in Clinical Texts} 
% \tnotetext[1]{You can use this document as the template for preparing your
%   publication. We recommend using the latest version of the ceurart style.}
\cortext[1]{Corresponding author.}
\fntext[1]{These authors contributed equally.}\author[1]{Huu-Huy-Hoang Tran}[%
email=23020073@vnu.edu.vn,
]
\fnmark[1]

\address[1]{University of Engineering and Technology, Vietnam National University}

\author[1]{Gia-Bao Duong}[%
email=23021475@vnu.edu.vn,
]
\fnmark[1]

\author[1]{Quoc-Viet-Anh Tran}[%
email=23021471@vnu.edu.vn,
]
\fnmark[1]

\author[1]{Thi-Hai-Yen Vuong}[%
email=yenvth@vnu.edu.vn,
]

\author[1]{Hoang-Quynh Le}[%
email=lhquynh@vnu.edu.vn,
]
\cormark[1]
\begin{abstract}
Extracting drug use information from unstructured Electronic Health Records (EHRs) remains a major challenge in clinical Natural Language Processing (NLP). While Large Language Models (LLMs) demonstrate advancements, their use in clinical NLP is limited by concerns over trust, control, and efficiency. To address this, we present NOWJ’s submission to the ToxHabits Shared Task at BioCreative IX (IJCAI 2025). This task targets the detection of toxic substance use and contextual attributes in Spanish clinical texts, a domain-specific, low-resource setting. We propose a multi-output ensemble system tackling both Subtask 1 (ToxNER) and Subtask 2 (ToxUse). Our system integrates BETO with a CRF layer for sequence labeling, employs diverse training strategies, and uses sentence filtering to boost precision. Our top run achieved 0.94 F1 and 0.97 precision for Trigger Detection, and 0.91 F1 for Argument Detection.
\end{abstract}


\maketitle

\section{Introduction}
In healthcare, it is crucial to effectively extract and use important information from unstructured Electronic Health Records (EHRs). Specifically, precisely extracting information on toxic substance use, as targeted by the ToxHabits Shared Task, is vital for improving patient management, supporting clinical research, and enhancing public health surveillance. To address this challenge, natural language processing (NLP) techniques can effectively identify, extract, and characterize specific mentions of substance use and their contextual attributes from these unstructured clinical texts.

We present NOWJ's submission to the ToxHabits Shared Task at the BioCreative IX Workshop (IJCAI 2025) \cite{toxhabitsoverview}, which focuses on detecting toxic substance use and its contextual attributes in Spanish clinical texts — a low-resource and domain-specific setting. The ToxHabits dataset \cite{toxhabitsDataset2025} consists of 1499 Spanish clinical case reports drawn from various open-access medical journals. These documents are annotated for Triggers (Tobacco, Cannabis, Alcohol, or Drug) and contextual Arguments (Type, Method, Amount, Frequency, Duration, History). Subtask 1 (ToxNER) focuses on identifying and classifying Trigger spans in Spanish clinical notes, while Subtask 2 (ToxUse) requires extracting and typing Argument spans. Performance for both subtasks is evaluated using micro-averaged precision, recall, and F1-scores \cite{ToxHabitsEval2025}. A detailed example of how the text was annotated is presented in Appendix~\ref{appendix:example-annotated-text}.

Although Large Language Models (LLMs) have achieved remarkable advancements across various NLP domains, their application in critical clinical settings often faces challenges related to trustworthiness and overall efficiency. In this report, we introduce a novel multi-output ensemble system designed to address both ToxNER and ToxUse simultaneously and achieve promising results.

Recent clinical NER efforts include the MultiCardio NER task \cite{MultiCardioNER}, which evaluated disease (DisTEMIST) and drug (DrugTEMIST) extraction in cardiology reports across Spanish, English, and Italian. The top approach combined an ensemble of RoBERTa \cite{liu2019roberta} models with a multi-head CRF, achieving an F1 of 0.82 on disease NER. The 2022 n2c2/UW shared task \cite{n2c2} focused on social determinants of health in English narratives; the winning system framed extraction as sequence-to-sequence with a T5 \cite{raffel2020exploring} encoder–decoder, reaching an F1 of 0.90.

To address the linguistic challenges inherent in Spanish clinical text, several research teams have effectively leveraged BETO \cite{CaneteCFP2020}, a BERT \cite{bert} model trained on a large Spanish corpus. This allows BETO to capture Spanish-specific morphology and syntax more effectively than multilingual models, making it well-suited for domain-specific tasks like clinical NER. As a result, BETO consistently outperforms general multilingual models in Spanish NLP tasks.

\section{Methodology}

We frame both subtasks as a single BIO‐based \cite{lafferty2001conditional} token classification problem. First, each document is split into sentences and passed through a binary filter that discards those without any potential triggers/arguments. Remaining sentences are tokenized and labeled under the BIO scheme. To combat overfitting on our small, domain‐specific dataset, we generate multiple resampled training subsets and train an ensemble of multi‐output BERT–CRF models, each jointly predicting Trigger and Argument tags. After inference, BIO tags are merged into complete entity spans. Finally, we aggregate model outputs via majority voting to enhance robustness and overall F1 performance across both subtasks.

\subsection{System Architecture}
\begin{figure}[h]
    \centering
    \includegraphics[width=1\textwidth]{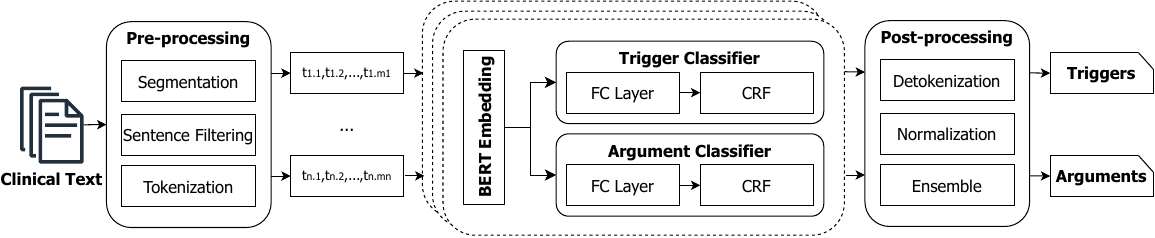}
    \caption{Overall architecture of our multi-output BERT-CRF ensemble system.}
    \label{fig:example}
\end{figure}

The proposed system is designed as a multi-stage neural network architecture for identifying triggers and arguments. The overall architecture of the system, described in Figure \ref{fig:example}, comprises three sequential components: Sentence pre-processing (segmentation, filtering, tokenization), Multi-output BERT-CRF modeling \cite{lafferty2001conditional}, and Detokenization and Ensemble inference. Each component is described in detail below.
\subsubsection{Pre-processing with Sentence Segmentation, Sentence Filtering, and Tokenization.}
To reduce computational burden, input clinical documents are first segmented into sentences. These sentences are then served as the input for the subsequent Sentence Filtering stage.

These sentences are then processed by a binary BERT-based classifier, which determines whether each sentence contains at least one relevant trigger or argument. This model utilizes a BETO backbone \cite{CaneteCFP2020} (the same as our main model) and was independently trained for this specific purpose. Specifically, we fine-tuned the BETO model on a subset of the ToxHabits training data, using sentences containing triggers or arguments as positive examples and sentences without any as negative examples.
The model takes a sentence as input and outputs a binary classification indicating its trigger/argument presence. In inference phase, only positively classified sentences are processed in subsequent stages.

The filtered sentences are then tokenized using the BETO model. Each token will later be labeled using a tagging scheme, independently for both tasks. This labeling strategy captures span boundaries by assigning B- (Beginning), I- (Inside), and O (Outside) tags to tokens based on whether they are part of a trigger or an argument.
\subsubsection{Multi-output Sequence Labeling with BERT-CRF.}
The core of our architecture is a multi-output sequence labeling model built upon a shared BETO encoder and task-specific decoding branches for trigger detection and argument identification. Each task branch applies a linear classification layer followed by a Conditional Random Field (CRF) \cite{lafferty2001conditional} decoder to generate the final label sequence. The inference process is structured as follows.\\
(1) Shared Encoder.

Given an input sentence tokenized into subword units \( (t_1, t_2, ..., t_n) \), we obtain contextual embeddings via the shared BETO encoder:
\[
\mathbf{H} = (h_1, h_2, ..., h_n), \quad h_i \in \mathbb{R}^d
\]
where \( d \) is the hidden size of BETO. These embeddings are passed through a dropout layer during training, resulting in \( \tilde{\mathbf{H}} \).
\\
(2) Task-specific Emission Layers.

Each branch projects \( \tilde{\mathbf{H}}\) into a task-specific label space via a fully connected layer:
\begin{itemize}
    \item Trigger detection branch:
    \[
    \mathbf{E}^{(\text{tr})} = \text{Linear}_{\text{tr}}(\tilde{\mathbf{H}}) \in \mathbb{R}^{n \times C_{\text{tr}}}
    \]
    \item Argument detection branch:
    \[
    \mathbf{E}^{(\text{arg})} = \text{Linear}_{\text{arg}}(\tilde{\mathbf{H}}) \in \mathbb{R}^{n \times C_{\text{arg}}}
    \]
\end{itemize}
where \( C_{\text{tr}} \) and \( C_{\text{arg}} \) denote the number of BIO labels for trigger and argument tagging, respectively. The emission vectors \( \mathbf{E} \) represent per-token label scores (logits).
\\
(3) CRF Decoding and Inference. 

To decode the most likely label sequence \( \hat{\mathbf{y}} \), we apply a CRF layer on top of the emission scores. The CRF models the conditional probability of a label sequence \( \mathbf{y} = (y_1, \dots, y_n) \) given the emission matrix \( \mathbf{E} \) as:
\[
P(\mathbf{y} \mid \mathbf{E}) = \frac{1}{Z(\mathbf{E})} \exp\left( \sum_{i=1}^{n} s_i(y_i) + \sum_{i=1}^{n-1} T_{y_i, y_{i+1}} \right)
\]
where:
\begin{itemize}
    \item \( s_i(y_i) = \mathbf{E}_{i, y_i} \) is the emission score of assigning label \( y_i \) to token \( i \),
    \item \( T_{y_i, y_{i+1}} \) is the trainable transition score between adjacent labels,
    \item \( \mathcal{Y} = Y^n \) denotes the set of all possible label sequences of length \( n \), with \( Y \) being the label vocabulary (e.g., BIO tags),
    \item \( Z(\mathbf{E}) \) is the partition function over \( \mathcal{Y} \):
    \[
    Z(\mathbf{E}) = \sum_{\mathbf{y'} \in \mathcal{Y}} \exp\left( \sum_{i=1}^{n} s_i(y'_i) + \sum_{i=1}^{n-1} T_{y'_i, y'_{i+1}} \right)
    \]
\end{itemize}

During inference, we decode the most probable label sequence \( \hat{\mathbf{y}} \) using the Viterbi algorithm:

\[
\hat{\mathbf{y}} = \arg\max_{\mathbf{y} \in \mathcal{Y}} P(\mathbf{y} \mid \mathbf{E})
\]

This decoding ensures global optimality under the transition constraints and returns valid BIO-structured outputs for each task.
\subsubsection{Post-processing with Detokenization, Normalization, and Ensemble Inference.}

The predictions at the subword level are reassembled and normalized into word-level spans. To enhance robustness, we aggregate outputs from multiple BERT-CRF models trained on different training strategies. The final output is derived through majority voting, enabling improved precision and reduced variance.
\subsection{Multiple Training Strategies for Ensemble}
For model optimization, we computed separate loss values for the two tasks—trigger detection and argument extraction—using their respective CRF layers. Specifically, the loss for trigger detection ($\mathcal{L}_{\text{tr}}$) and the loss for argument extraction ($\mathcal{L}_{\text{arg}}$) were both calculated based on the negative log-likelihood from each CRF. The overall training objective was the simple sum of these two components:

\begin{equation}
    \mathcal{L} = \alpha*\mathcal{L}_{\text{tr}} + \beta*\mathcal{L}_{\text{arg}}
\end{equation}
where $\alpha$ and $\beta$ are the weights, both set to 1 in this work.

This joint loss formulation allowed the model to learn both tasks simultaneously, balancing shared feature extraction from the encoder with task-specific predictions from the CRF layers.

To improve model generalization and address class imbalance, we applied several data preparation and training strategies. First, we utilized two training datasets: the partial training dataset and the full training dataset. To increase data diversity, we split each training dataset into $k$ equal parts. For each split, we trained a model on $k-1$ parts, leaving out a different part in each iteration. This procedure resulted in $k$ distinct models per dataset. For model training, we experimented with three strategies:
\begin{itemize}
    \item \textbf{Label-Weighted Loss:} Assigning higher weights to underrepresented classes in the loss function to mitigate class imbalance during optimization.
    \item \textbf{Data Oversampling:} Increasing the number of training instances containing triggers or arguments by duplicating such sentences in the training data to balance the class distribution.
    \item \textbf{Weighted Random Sampling:} Increasing the sampling probability of sentences containing triggers/arguments to ensure that the model is exposed to more informative samples during each training epoch.
\end{itemize} 
By combining these strategies with the resampled datasets, we created multiple training configurations to ensemble different variants of the multi-output BERT-CRF model. A total of $n$ models were selected and used for both trigger and argument detection. These word-level outputs were subsequently aggregated across models via majority voting to produce the final predictions for triggers and arguments.

\section{Experiments and Results}

\subsection{Dataset Analysis}
The ToxHabits corpus \cite{toxhabitsDataset2025} comprises 1,499 Spanish clinical case reports, split into training and test sets as shown in Table~\ref{tab:dataset_stats}. Each training document includes a text file and two annotation files, one per subtask, with aligned trigger and argument annotations in .ann format. Each annotation specifies the filename, label, start and end offsets, and span text. There are four trigger types (Tobacco, Cannabis, Alcohol, Drug) and six argument types (Type, Method, Amount, Frequency, Duration, History).

\begin{table}[h]
    \centering
    \begin{tabular}{lcc}
         \toprule
         \textbf{Metric} & \textbf{Train} & \textbf{Test (private)} \\
         \midrule
         \ Number of Documents & 1199 & 300 \\
         \ Average Document Length (words) & 516 & 498 \\
         \ Number of Triggers & 7592 & - \\
         \ Number of Arguments & 9528 & - \\
         \ Average number of Triggers per Document & 6.33 & - \\
         \ Average number of Arguments per Document & 7.95 & - \\
        \bottomrule  
    \end{tabular}
    \caption{Descriptive statistics of the ToxHabits dataset, including the number of documents, average document length, and average number of annotated triggers and arguments per document in the training set.}
    \label{tab:dataset_stats}
\end{table}

To analyze dataset characteristics, we performed a detailed count of annotated triggers and arguments. The training set has 1,199 documents (avg. 516 words/doc), and the test set 300 (avg. 498 words/doc), showing consistent lengths. There are 7,592 triggers and 9,528 arguments in training, averaging 6.33 triggers and 7.95 arguments per document. This suggests frequent mentions of substance use, with arguments often exceeding trigger counts per document.

\subsection{Evaluation Metrics}
For both ToxNER (Subtask 1) and ToxUse (Subtask 2), our systems were officially evaluated by the ToxHabits Shared Task Organizer, according to the official evaluation guidelines \cite{ToxHabitsEval2025}. The evaluation compared our system's generated entities with the expert-annotated ground truth dataset, considering predictions correct only if they exactly matched the gold standard in both character span and label. The primary evaluation metrics were micro-averaged precision, recall, and F1-scores.

\subsection{Experimental Settings}
Our core model architecture is BETO \cite{CaneteCFP2020} in its "base" size variant, with a Conditional Random Field (CRF) layer for sequence labeling. All implementations are based on the Hugging Face Transformers library with PyTorch, and we utilized two NVIDIA Tesla T4 GPUs for our experiments. For fine-tuning the main BERT-CRF model, we adhered to specific configurations. Detailed hyperparameters for this model are provided in Appendix \ref{appendix:main-model-details} for full reproducibility. Separately, the binary BERT-based classifier used for sentence filtering was trained independently. Its specific hyperparameters and training configurations are detailed in Appendix \ref{appendix:sentence-filtering-model-details}.

To enhance robustness across different scenarios, we employed two distinct ensemble configurations. For the Full Training Dataset Evaluation, an ensemble of 6 models was trained using Weighted Random Sampling on data from a 5-fold cross-validation split (5 models on 80\% subsets and 1 on the 100\% full dataset). For the Partial Training Dataset Evaluation, we constructed a larger ensemble of 19 models. These models leveraged diverse data samples and combined three training strategies: Label-Weighted Loss (6 models: 5 on 80\% resampled, 1 on full partial), Weighted Random Sampling (12 models: 10 on 80\% resampled, 2 on full partial from two runs), and Data Oversampling (2 models, one with specific ratios 9-3-2-2 for Drug, Alcohol, Tobacco, Cannabis, and another with a uniform 1-1-1-1 distribution).
\subsection{Experimental Results}
We conducted five official runs with distinct configurations, encompassing:
(i) training with either the full or partial dataset,
(ii) applying our developed sentence filtering method to remove sentences irrelevant to trigger or argument information, and
(iii) fine-tuning model parameters.

% Requires: \usepackage{graphicx}
\subsubsection{Subtask 1: Trigger Detection (ToxNER)}
\begin{table}[h]
    \centering
    \label{tab:results-subtask1}
    \begin{tabular}{lccc}
        \hline
        \textbf{Model / Method (Ensemble)} & \textbf{Precision} & \textbf{Recall} & \textbf{F1-score} \\
        \hline
        BERT Ensemble (full train) & 0.92 & \textbf{0.95} & 0.93 \\
        BERT Ensemble + Sentence Filtering (full train) & 0.94 & 0.94 & \textbf{0.94} \\
        BERT Ensemble (partial train) & 0.83 & 0.85 & 0.84 \\
        BERT Ensemble + Sentence Filtering (partial train) & 0.83 & 0.85 & 0.84 \\
        BERT Ensemble + Sentence Filtering + Tuning (full train) & \textbf{0.97} & 0.92 & \textbf{0.94} \\
        \hline
    \end{tabular}
    \caption{Results on the test set of Subtask 1 (ToxNER).}
\end{table}

The official results for Subtask 1, Trigger Detection (ToxNER), are summarized in Table \ref{tab:results-subtask1}. As shown in the Table, the best-performing run for Trigger Detection was "BERT Ensemble + Sentence Filtering + Tuning (full train)", achieving an F1-score of 0.94 and a precision of 0.97. The "BERT Ensemble + Sentence Filtering (full train)" model also achieved an F1-score of 0.94, with a precision of 0.94. Notably, the sentence filtering method improved the F1-score by 0.01 points (from 0.93 to 0.94) when applied to the full training dataset. Models trained on the full dataset consistently outperformed those that only took advantage of the partial dataset.

\subsubsection{Subtask 2: Argument Detection (ToxUse)}
For Argument Detection, we adopted the same modeling architecture and configuration design as in Subtask 1, as our methodology uses a multi-output approach for both subtasks. The official results for Subtask 2, Argument Detection (ToxUse), are summarized in Table \ref{tab:results-subtask2}.

\begin{table}[h]
    \centering
    \label{tab:results-subtask2}
    \begin{tabular}{lccc}
        \hline
        \textbf{Model / Method (Ensemble)} & \textbf{Precision} & \textbf{Recall} & \textbf{F1-score} \\
        \hline
        BERT Ensemble (full train) & 0.91 & \textbf{0.90} & \textbf{0.91} \\
        BERT Ensemble + Sentence Filtering (full train) & 0.92 & 0.88 & 0.90 \\
        BERT Ensemble (partial train) & 0.78 & 0.75 & 0.77 \\
        BERT Ensemble + Sentence Filtering (partial train) & 0.79 & 0.75 & 0.77 \\
        BERT Ensemble + Sentence Filtering + Tuning (full train) & \textbf{0.95} & 0.85 & 0.90 \\
        \hline
    \end{tabular}
    \caption{Results on the test set of Subtask 2 (ToxUse).}
\end{table}

As presented in Table \ref{tab:results-subtask2}, the highest F1-score of 0.91 for Argument Detection was achieved by the "BERT Ensemble (full train)" model. Notably, the "BERT Ensemble + Sentence Filtering + Tuning (full train)" model recorded the highest precision at 0.95. While the sentence filtering method for Argument Detection generally increased precision, it also led to a slight decrease in recall, resulting in a marginal F1-score reduction compared to the base full-train model (from 0.91 to 0.90).

\subsection{Discussion}
Our results highlight the effectiveness of our multi-output ensemble approach, which achieved promising performance for both Trigger and Argument Detection. The multi-output architecture specifically allowed the model to learn shared representations and dependencies between triggers and arguments. This joint learning, along with our ensemble method, significantly enhanced the system's robustness and generalization. In our experiments, we chose to set both $\alpha$ and $\beta$ weights in the joint loss function to 1 for simplicity and to ensure equal emphasis on trigger detection and argument extraction. However, when the frequency of triggers and arguments differs substantially, adjusting these weights could be a potential option for further improvements. The sentence filtering method performed efficiently across most runs, consistently increasing precision and boosting F1-scores in most configurations. Currently, our system does not utilize Large Language Models (LLMs); however, we consider exploring their integration in the future to advance generalization and contextual understanding further.

\section{Conclusion}

In summary, this report introduced a novel multi-output ensemble system for simultaneous Trigger (ToxNER) and Argument (ToxUse) detection. Our multi-stage methodology combined initial sentence filtering with a multi-output BERT-CRF core and an ensemble inference strategy. Evaluated on the official ToxHabits Shared Task datasets, our approach demonstrated competitive performance. We hope our methods and experiences will prove valuable for future participants and research in clinical information extraction.

% \begin{acknowledgments}
% We thank the organizers of the BioCreative IX ToxHabits Shared Task for providing the dataset, evaluation framework, and continuous support throughout the competition. We also express our gratitude to the Data Science and Knowledge Technology Laboratory (DS\&KT Lab) at VNU University of Engineering and Technology for their guidance and computational resources. Lastly, we thank all team members for their dedication and collaboration during the development of our system.
% \end{acknowledgments}

% %% The declaration on generative AI comes in effect
% %% in Janary 2025. See also
% %% https://ceur-ws.org/GenAI/Policy.html
% \section*{Declaration on Generative AI}
% During the preparation of this work, the author(s) employed ChatGPT-4o and Gemini 2.5 to assist with improving writing style, grammar and spelling checks, and paraphrasing and rewording. No generative AI tools were used to produce images, figures, or data. After using these tools/services, the author(s) carefully reviewed and edited the content as needed and took full responsibility for the final manuscript.
%%
%% Define the bibliography file to be used
\bibliography{main-ref}

\appendix
\section{Example of annotated text}
\label{appendix:example-annotated-text}
\begin{figure}[h]
    \centering
    \includegraphics[width=0.8\textwidth]{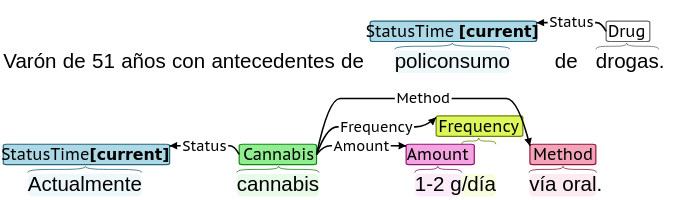}
    \caption{An example of an annotated clinical text snippet from the ToxHabits corpus (Image courtesy of ToxHabits Shared Task Organizers \cite{toxhabitsoverview})}
    \label{fig:annotated_example}
\end{figure}

To illustrate the annotation scheme and the types of entities targeted, consider the example from the ToxHabits corpus shown in Figure \ref{fig:annotated_example}. This figure, provided by the ToxHabits Shared Task organizers, clarifies how Trigger and Argument spans are marked within the clinical text. 

For instance, in the Spanish sentences: "Varón de 51 años con antecedentes de policonsumo de drogas. Actualmente, cannabis 1-2 g/día vía oral." (Translation: "51-year-old man with past history of multi-drug consumption. Nowadays, (he takes) 1-2 g/day of cannabis orally."). Here, "drogas" is annotated as a Trigger: Drug. Additionally, "cannabis" is identified as a Trigger: Cannabis, with "1-2 g" serving as an Argument: Amount, "/día" as an Argument: Frequency, and "vía oral" as an Argument: Method.

The figure also includes annotations of the StatusTime label; however, this label is not used for identification and evaluation in either of the two subtasks considered in our system.

\section{Main model Configurations}
\label{appendix:main-model-details}
This section details the training configurations and hyperparameters for the main multi-output BERT-CRF model. These parameters, crucial for the model's performance and reproducibility, are itemized in Table~\ref{tab:bertcrf-config}. The total training time was approximately 1 hour.

\begin{table}[h]
    \centering
    \begin{tabular}{ll}
        \hline
        \textbf{Hyperparameter}           & \textbf{Value}    \\
        \hline
        Max length                    & 512                \\
        Optimizer                     & AdamW              \\
        Learning rate (LR)            & 5e-5               \\
        Batch size                    & 8                  \\
        Dropout                       & 0.1                \\
        Epochs                        & 5                  \\
        \hline
    \end{tabular}
    \caption{BERT-CRF fine-tune configurations}
    \label{tab:bertcrf-config}
\end{table}

\section{Sentence Filtering Model Configurations}
\label{appendix:sentence-filtering-model-details}
The training configuration and hyperparameters for the sentence filtering model are detailed in Table~\ref{tab:sentence-filtering-config}. The model was trained using the CrossEntropy loss function, and the total training time was approximately 2.5 hours.

\begin{table}[h]
    \centering
    \begin{tabular}{ll}
        \hline
        \textbf{Hyperparameter}           & \textbf{Value}    \\
        \hline
        Max length                    & 512                \\
        Optimizer                     & AdamW              \\
        Learning rate (LR)            & 1e-5               \\
        Batch size                    & 16                 \\
        Dropout                       & 0.1                \\
        Epochs                        & 5                  \\
        Weight decay                  & 0.01               \\
        \hline
    \end{tabular}
    \caption{BERT-based Sentence Filtering fine-tune configurations}
    \label{tab:sentence-filtering-config}
\end{table}

\end{document}